\documentclass{article}

\PassOptionsToPackage{numbers, compress}{natbib}

\usepackage[preprint]{neurips_2023}

\usepackage[utf8]{inputenc} 
\usepackage[T1]{fontenc}    
\usepackage{hyperref}       
\usepackage{url}            
\usepackage{booktabs}       
\usepackage{amsfonts}       
\usepackage{nicefrac}       
\usepackage{microtype}      
\usepackage{xcolor}         

\usepackage{amsmath}
\usepackage{derivative}
\usepackage{caption}
\usepackage{subcaption}
\usepackage{graphicx}
\usepackage{pifont}  
\usepackage{adjustbox}  
\usepackage{multirow}

\DeclareMathOperator{\PatchEmbedding}{PatchEmbedding}
\DeclareMathOperator{\pos}{pos}

\DeclareMathOperator{\ReLU}{ReLU}
\DeclareMathOperator{\Attention}{Attention}
\DeclareMathOperator{\softmax}{softmax}
\DeclareMathOperator{\rel}{rel}
\DeclareMathOperator{\Gaussian}{Gaussian}

\newcommand*{\RS}{\ensuremath{R^2}}
\newcommand*{\hsX}{\ensuremath{\hat{\sigma_X}}}
\newcommand*{\hsY}{\ensuremath{\hat{\sigma_Y}}}
\newcommand*{\real}{\ensuremath{\mathbb{R}}}
\newcommand*{\Brl}{\ensuremath{\mathbf{B}_{\rel,l}}}
\newcommand*{\BGl}{\ensuremath{\mathbf{B}_{\Gaussian,l}}}

\def\figref#1{Figure~\ref{#1}}
\def\tabref#1{Table~\ref{#1}}
\def\eqref#1{Eq.~\ref{#1}}

\title{Understanding Gaussian Attention Bias of Vision Transformers Using Effective Receptive Fields}

\author{%
    Bum Jun Kim\\
    POSTECH\\
  \texttt{kmbmjn@postech.edu} \\
  \And
  Hyeyeon Choi\\
    POSTECH\\
  \texttt{hyeyeon@postech.edu} \\
  \And
    Hyeonah Jang\\
    POSTECH\\
  \texttt{hajang@postech.edu} \\
  \And
    Sang Woo Kim\\
    POSTECH\\
  \texttt{swkim@postech.edu} \\
}

\begin{document}

\maketitle

\begin{abstract}
	Vision transformers (ViTs) that model an image as a sequence of partitioned patches have shown notable performance in diverse vision tasks. Because partitioning patches eliminates the image structure, to reflect the order of patches, ViTs utilize an explicit component called positional embedding. However, we claim that the use of positional embedding does not simply guarantee the order-awareness of ViT. To support this claim, we analyze the actual behavior of ViTs using an effective receptive field. We demonstrate that during training, ViT acquires an understanding of patch order from the positional embedding that is trained to be a specific pattern. Based on this observation, we propose explicitly adding a Gaussian attention bias that guides the positional embedding to have the corresponding pattern from the beginning of training. We evaluated the influence of Gaussian attention bias on the performance of ViTs in several image classification, object detection, and semantic segmentation experiments. The results showed that proposed method not only facilitates ViTs to understand images but also boosts their performance on various datasets, including ImageNet, COCO 2017, and ADE20K.
\end{abstract}

\section{Introduction}
\label{sec:int}
Vision transformers (ViTs) \citep{DBLP:conf/iclr/DosovitskiyB0WZ21} have achieved remarkable performances in various vision tasks that are often superior to those of convolutional neural networks (CNNs) \citep{DBLP:conf/cvpr/ZhengLZZLWFFXT021,DBLP:conf/nips/XieWYAAL21,DBLP:conf/eccv/CarionMSUKZ20}. Unlike CNNs, ViTs partition an image into a sequence of patches and subsequently combine patch features based on the self-attention (SA) mechanism, enabling the aggregation of rich global information within the image \citep{DBLP:conf/iclr/CordonnierLJ20,DBLP:conf/cvpr/ZhaoJK20,DBLP:conf/nips/KossenBLGRG21}.

Despite its effectiveness, SA poses inherent limitations in understanding the order of input patches. However, because 2D images are structured data, understanding the order of patches is important for ViT \citep{DBLP:conf/iclr/Bello21}. To overcome this problem, ViTs employ an explicit component called positional embedding that enables the identification of the order and the corresponding geometric positions of patches.

However, we claim that simply using positional embeddings does not ensure order-awareness. To validate our claim, we utilize the effective receptive field (ERF) \citep{DBLP:conf/nips/LuoLUZ16,araujo2019computing} that highlights the pixels actually used in perception, illustrating how ViTs understand images. Using ERF, we demonstrate that ViTs with untrained positional embeddings do not discriminate between near and far patches and that order-awareness is acquired after positional embedding is trained to obtain specific patterns.

Motivated by this observation, to construct a ViT born with the spatial understanding of images, we propose injecting Gaussian attention bias into positional embedding. The innate spatial understanding of images helps ViT capture the nearness and farness of pixels, thereby enhancing the performance of ViT in vision tasks. We observed that using Gaussian attention bias improved the performance of ViTs on several datasets, tasks, and models.

\section{Background}
\label{sec:bac}
\paragraph{Forward propagation of standard ViT} Let $\mathbf{x} \in \real^{H \times W \times C}$ be the input image to the ViT, where $H \times W$ is the resolution, and $C$ is the number of channels of the image. Using a predefined resolution of patch $P \times P$, ViT spatially partitions the image into $N$ nonoverlapping patches, where $N=HW/P^2$. Each patch is linearly projected, yielding $\PatchEmbedding(\mathbf{x})$.\footnote{The concatenation of the class token is ignored in this study for notational simplicity.} Subsequently, an absolute positional embedding $\mathbf{E}_{\pos}$ is added, resulting in $\mathbf{z}_0$, the input to the first transformer block. Now, transformer blocks containing SAs and multilayer perceptrons (MLPs) are applied in a row to produce $\mathbf{z}_L$, where $L$ is the number of transformer blocks. Finally, LayerNorm \citep{DBLP:journals/corr/BaKH16} is applied to produce the last feature map $\mathbf{y}$ from which the head produces a classification score. Here, $\PatchEmbedding(\mathbf{x})$, $\mathbf{E}_{\pos}$, $\mathbf{z}_l$, and $\mathbf{y}$ have the same size of $\real^{N\times D}$, where $D$ is the dimension of the patch features.

\paragraph{Trick to obtain the ERF of ViT} Because the ERF of ViT has been rarely discussed and our study is the first to provide a concrete and detailed analysis on this topic, we first formulate the ERF of ViT. The ERF depicts the actual usage of each pixel for determining the target feature in a neural network, representing a generic connection between them. We follow the common tricks to obtain ERFs of CNNs \citep{DBLP:journals/prl/KimCJLJK23}. However, unlike CNNs, to obtain the ERF of ViT, we should focus on the patch unit. To investigate the properties of the entire ViT, the last feature map $\mathbf{y}$ is chosen as the target feature map. First, we target the $n$th patch corresponding to the central patch. Because our goal is to analyze the spatial relationship between the target patch and pixel units, we ignore other units such as the image channel and the dimension of the patch feature. Thus, the features of the central patch are averaged over its dimensions: $Y=\frac{1}{D}\sum_{d=1}^{D} {\mathbf{y}_{n,d}}$. Now, we examine the contribution of each pixel to $Y$ that can be obtained by a gradient $\pdv{Y}{\mathbf{x}} \in \real^{H \times W \times C}$. The gradient is averaged over channels to obtain $\mathbf{G}=\frac{1}{C} \sum_{c=1}^{C} {[\pdv{Y}{\mathbf{x}}]_c}$. At this point, $\mathbf{G} \in \real^{H \times W}$ contains the spatial relationship between the targeted patch feature and pixels. However, because $\mathbf{G}$ arises from the forward and backward operations of a single image, $\mathbf{G}$ is strongly dependent on the input image rather than on the ViT's properties. To capture the general behavior of the ViT, $\mathbf{G}$ is averaged over a sufficiently large number of images. At this time, because negative values in $\mathbf{G}$ cancel out the positive values, we ignore the negative importance using $\ReLU$ \citep{DBLP:journals/prl/KimCJLJK23,DBLP:journals/ijcv/SelvarajuCDVPB20,DBLP:conf/wacv/ChattopadhyaySH18}. Thus, we obtain $\mathbf{R}=\frac{1}{\lvert S \rvert} \sum_{\mathbf{x} \in S} \ReLU(\mathbf{G})$, where $S$ denotes an image dataset. Because it is averaged over numerous images, $\mathbf{R} \in \real^{H \times W}$ represents the general relationship between the pixels and the targeted patch feature of the ViT corresponding to ERF.

\section{Effective Receptive Fields of Vision Transformers}
\label{sec:eff}
\subsection{Qualitative Analysis}
\label{subsec:analysis}
\begin{figure}[t!]
    \begin{center}
        \begin{tabular}{cccc}
            \includegraphics[width=0.24\textwidth]{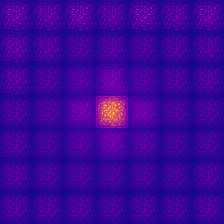} &
            \includegraphics[width=0.24\textwidth]{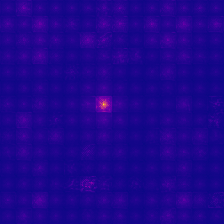} &
            \includegraphics[width=0.24\textwidth]{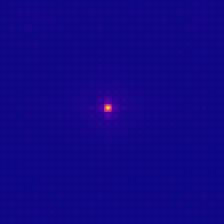} &
            \includegraphics[width=0.24\textwidth]{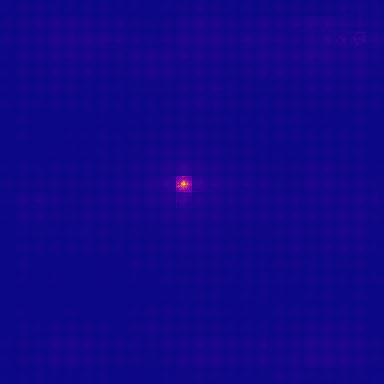} \\
            (a) ViT-B/32, 224$^2$ & (b) ViT-B/16, 224$^2$ & (c) ViT-B/8, 224$^2$ & (d) ViT-B/16, 384$^2$
        \end{tabular}
    \end{center}
    \caption{ERFs of ViT with different patch sizes. Because printed figures can be seen improperly, we highly encourage viewing all images electronically with zoom.}
    \label{fig:patchsize}
\end{figure}

In this section, we qualitatively analyze the ERFs of ViTs. Although the ERFs of ResNets \citep{DBLP:conf/cvpr/HeZRS16,DBLP:conf/cvpr/XieGDTH17} resemble a 2D Gaussian, the ERFs of ViTs exhibit a different shape owing to nonoverlapping patch partitioning \citep{DBLP:conf/nips/RaghuUKZD21}. \figref{fig:patchsize} shows the ERFs of ViT-B with different patch sizes of $\{32, 16, 8\}$. First, the ERF of ViT mainly highlights the targeted central patch and slightly uses information from other patches. This behavior indicates that each patch feature is responsible for representing information in the corresponding patch, while it is combined with certain global information from other patches via SA.

\begin{figure}[t!]
	\begin{center}
		\begin{tabular}{cccc}
            \includegraphics[width=0.24\textwidth]{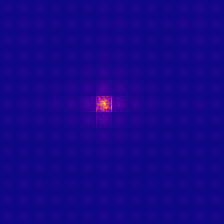} &
			\includegraphics[width=0.24\textwidth]{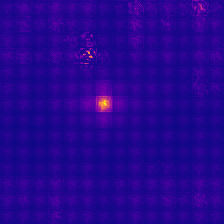} &
			\includegraphics[width=0.24\textwidth]{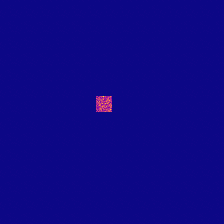} &
			\includegraphics[width=0.24\textwidth]{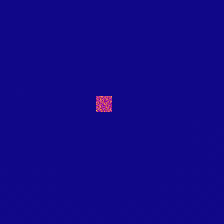} \\
			(a) ViT-S/16 & (b) ViT-L/16 & (c) ViT-S/16 (U) & (d) ViT-L/16 (U)
		\end{tabular}
	\end{center}
    \caption{ERFs of ViT with different model sizes. (U) indicates an untrained model.}
	\label{fig:modelsize}
\end{figure}


For ViT, a large-sized model was observed to yield a widespread ERF (\figref{fig:patchsize} (b) and \figref{fig:modelsize} (a, b)). This observation is expected because a larger ViT further stacks wider layers. However, for untrained ViTs, the ERFs exhibited no difference (\figref{fig:modelsize} (c, d)). This observation shows that the wider ERF of the ViT-L/16 arises from not only architectural largeness but also pretrained weights.

\begin{figure}[t!]
	\begin{center}
		\begin{tabular}{cccc}
            \includegraphics[width=0.24\textwidth]{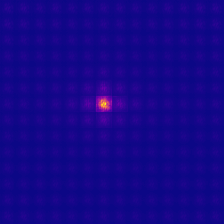} &
            \includegraphics[width=0.24\textwidth]{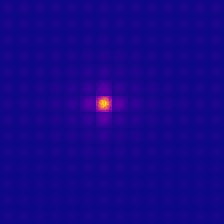} &
            \includegraphics[width=0.24\textwidth]{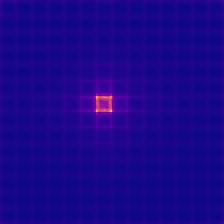} &
            \includegraphics[width=0.24\textwidth]{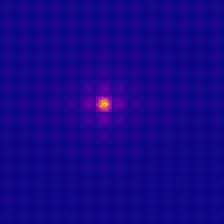} \\
			(a) DeiT-B/16 \citep{DBLP:conf/icml/TouvronCDMSJ21}&(b) DeiT III-B/16 \citep{DBLP:conf/eccv/TouvronCJ22}&(c) BEiT-B/16 \citep{DBLP:conf/iclr/Bao0PW22}&(d) CaiT-S-24 \citep{DBLP:conf/iccv/TouvronCSSJ21} \\ \\
            \includegraphics[width=0.24\textwidth]{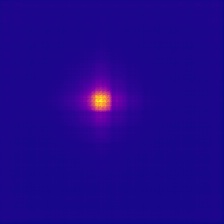} &
            \includegraphics[width=0.24\textwidth]{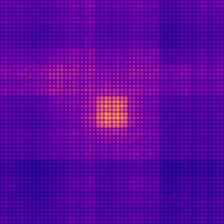} &
            \includegraphics[width=0.24\textwidth]{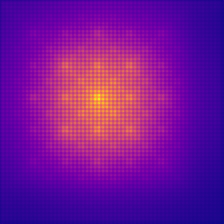} &
			\includegraphics[width=0.24\textwidth]{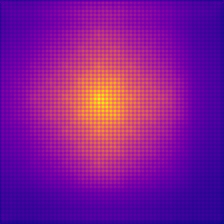} \\
			(e) XCiT-M24/16 \citep{DBLP:conf/nips/AliTCBDJLNSVJ21}&(f) Swin-B \citep{DBLP:conf/iccv/LiuL00W0LG21}&(g) ResNet-50 \citep{DBLP:conf/cvpr/HeZRS16}&(h) X-101 \citep{DBLP:conf/cvpr/XieGDTH17}
		\end{tabular}
	\end{center}
	\caption{ERFs of ViTs and CNNs. ``X'' indicates ResNeXt with 32 cardinality.}
	\label{fig:variousmodels}
\end{figure}

In addition, we obtained ERFs of other ViT variants (\figref{fig:variousmodels}). Among them, the DeiT \citep{DBLP:conf/icml/TouvronCDMSJ21}, DeiT III \citep{DBLP:conf/eccv/TouvronCJ22}, and BEiT \citep{DBLP:conf/iclr/Bao0PW22} have nearly identical architectures to the ViT with different pretrained weights, yielding similar but slightly different ERFs. CaiT \citep{DBLP:conf/iccv/TouvronCSSJ21} exhibits architectural modifications, such as class attention, but yielded a similar ERF to that of ViT. Compared with others, XCiT \citep{DBLP:conf/nips/AliTCBDJLNSVJ21} and Swin \citep{DBLP:conf/iccv/LiuL00W0LG21} exhibit wider ERFs. These include explicit modules that allow communication with the neighboring patches, such as the local patch interaction module, patch merging layer, and shifted window partitioning. In summary, we observed that the ERFs of ViTs were represented as highlights of the targeted patch area with other patches being partially utilized (See the Appendix for further analysis of the ERF of ViT).

\subsection{Spatial Understanding of Images by ViTs}
\label{subsec:spatial}
The interesting aspect of ERF is that it illustrates how ViTs understand spatial images. Although the ERF of ViT shows that the majority of the activated pixels are in the target patch, adjacent patches are more activated than distant patches, yielding a roughly \ding{58}-shape. This behavior implies that the ViTs have order-awareness in the sequence of patches, enabling them to use more information from nearby patches and less from far patches, which we refer to as the spatial understanding of images. The spatial understanding of images is one of the critical components for obtaining high-performance ViT. References \citep{DBLP:conf/iccv/WuPCFC21} and \citep{DBLP:conf/iccv/LiuL00W0LG21} observed that ViTs without any positional embedding yielded decreased performance.

Positional embeddings in ViTs exist in diverse forms. The absolute positional embedding (APE) can be either a predefined sinusoidal sequence \citep{DBLP:conf/nips/VaswaniSPUJGKP17} or learnable parameter \citep{DBLP:conf/iclr/DosovitskiyB0WZ21} and is added to the patch embedding. The relative positional embedding (RPE) \citep{DBLP:conf/naacl/ShawUV18,DBLP:conf/iclr/HuangVUSHSDHDE19}, also called attention bias \citep{DBLP:conf/iccv/GrahamETSJJD21}, is added to the attention matrix for each layer as
\begin{align}
    \Attention_l(\mathbf{Q}_l, \mathbf{K}_l, \mathbf{V}_l) = \softmax\left(\frac{\mathbf{Q}_l\mathbf{K}_l^\top}{\sqrt{D}} + \Brl\right) \mathbf{V}_l, \label{eq:sa}
\end{align}
where $\mathbf{Q}_l, \mathbf{K}_l, \mathbf{V}_l \in \real^{N \times D}$ are the query, key, and value in SA, respectively, and $\Brl \in \real^{N \times N}$ is the RPE as attention bias. To obtain $\Brl$, the original Swin transformer \cite{DBLP:conf/iccv/LiuL00W0LG21} used a learnable table called RelPosBias that provided $\Brl$ for each relative coordinate. SwinV2 \citep{DBLP:conf/cvpr/Liu0LYXWN000WG22} employed a learnable MLP to obtain $\Brl$ from each relative coordinate; this term is also called RelPosMlp. Although the original ViT used APE \citep{DBLP:conf/iclr/DosovitskiyB0WZ21}, recent ablation studies \citep{DBLP:conf/cvpr/SrinivasLPSAV21,DBLP:conf/iclr/Bao0PW22,DBLP:conf/iccv/WuPCFC21,DBLP:conf/iccv/LiuL00W0LG21} have reported that using RPE yielded improved performance. Reference \citep{DBLP:conf/iclr/Bello21} claims that although APE has provided successful results in natural language processing tasks, relative information from RPE is crucial for vision tasks.

\begin{figure}[t!]
	\begin{center}
		\begin{tabular}{cccc}
			\includegraphics[width=0.24\textwidth]{images/erf_dAdi_new_in_vit_small_patch16_224caittype_Tplasma.png} &
            \includegraphics[width=0.24\textwidth]{images/erf_dAdi_new_in_vit_base_patch16_224caittype_Tplasma.png} &
            \includegraphics[width=0.24\textwidth]{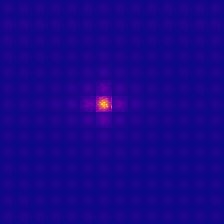} &
            \includegraphics[width=0.24\textwidth]{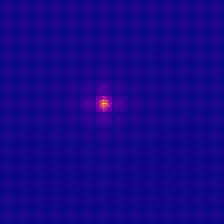} \\
            \includegraphics[width=0.24\textwidth]{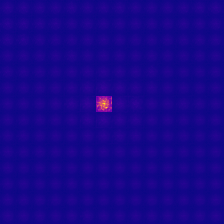} &
            \includegraphics[width=0.24\textwidth]{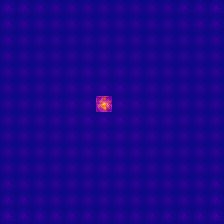} &
            \includegraphics[width=0.24\textwidth]{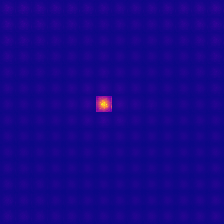} &
			\includegraphics[width=0.24\textwidth]{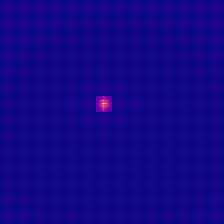} \\
			(a) ViT-S/16                                                                                                       & (b) ViT-B/16 & (c) ViT-M/16 (R) & (d) ViT-B/16 (R)
		\end{tabular}
	\end{center}
	\caption{ERFs of ViTs, where (R) indicates the model with RPE. The second row illustrates ERFs when the APE or RPE is re-initialized to random parameters. Note that the \ding{58}-shape is lost in the second row.}
	\label{fig:erfpos}
\end{figure}

We examined the role of positional embeddings. We obtained ERFs before and after the APE or RPE of pretrained ViTs was re-initialized to random parameters (\figref{fig:erfpos}). We observed that re-initializing APE or RPE altered ERFs, causing adjacent patches to lose their contribution to the target patch feature and exhibit the same contribution as far patches. This observation clearly demonstrates that SA itself cannot understand the location of patches, and positional embedding plays a significant role in the spatial understanding of images.

\begin{figure}[t!]
	\begin{center}
		\begin{tabular}{cccc}
            \includegraphics[width=0.24\textwidth]{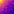} &
            \includegraphics[width=0.24\textwidth]{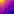} &
            \includegraphics[width=0.24\textwidth]{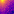} &
			\includegraphics[width=0.24\textwidth]{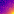} \\
			(a) $n=0$                                                                                                             & (b) $n=1$ & (c) $n=2$ & (d) $n=195$ \\ \\
            \includegraphics[width=0.24\textwidth]{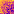} &
            \includegraphics[width=0.24\textwidth]{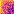} &
            \includegraphics[width=0.24\textwidth]{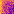} &
			\includegraphics[width=0.24\textwidth]{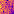} \\
			(e) $n=0$                                                                                                             & (f) $n=1$ & (g) $n=2$ & (h) $n=195$
		\end{tabular}
	\end{center}
	\caption{RPE of ViT-B/16 (R) for each patch index. The first row is obtained from the pretrained model, whereas the second row is obtained from the untrained model.}
	\label{fig:relposbypatch}
\end{figure}

To investigate the underlying mechanism, we extracted the learned and untrained RPEs. From $\Brl \in \real^{N \times N}$, the RPE of the $n$th patch $\mathbf{B}_{\rel, l, n} \in \real^{N}$ was obtained for $n \in \{0, 1, \cdots, N-1\}$ and was reshaped into $\mathbf{B}_{\rel, l, n}^{\prime} \in \real^{H/P \times W/P}$. For visualization, $\mathbf{B}_{\rel, l, n}^{\prime}$ was averaged over multi-head. The RPE in the first attention layer is visualized in \figref{fig:relposbypatch}. Note that the learned RPE appeared as a sliced 2D Gaussian, distinguishing between near and distant patches. Because $\softmax$ computes an exponential ratio, the bias term $\Brl$ in \eqref{eq:sa} becomes an exponential coefficient that amplifies each element of the attention matrix. Thus, a higher RPE value means a larger amplification of the corresponding patch. Importantly, we observed that untrained RPEs showed distance-independent values. In other words, re-initializing RelPosMlp provides a random RPE that does not discriminate between near and far patches.

\begin{figure}[t!]
	\begin{center}
		\begin{tabular}{cccc}
            \includegraphics[width=0.24\textwidth]{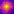} &
            \includegraphics[width=0.24\textwidth]{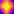} &
            \includegraphics[width=0.24\textwidth]{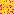} &
			\includegraphics[width=0.24\textwidth]{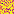} \\
			(a) $l=1$                                                                                                   & (b) $l=10$ & (c) $l=11$ & (d) $l=12$
		\end{tabular}
	\end{center}
	\caption{RPE corresponding to the center was extracted for each layer of ViT-B/16 (R).}
	\label{fig:relposbylayer}
\end{figure}


\begin{table}[t!]
	\begin{center}
		\adjustbox{max width=\textwidth}{
			\begin{tabular}{l|rrr|rrr|rrr}
				\toprule
				    & \multicolumn{3}{c|}{ViT-S/16, 224$^2$ (R)} & \multicolumn{3}{c|}{ViT-M/16, 224$^2$ (R)} & \multicolumn{3}{c}{ViT-B/16, 224$^2$ (R)}                                                         \\
				$l$ & $\RS$                                     & $\hsX$                                    & $\hsY$                                   & $\RS$ & $\hsX$  & $\hsY$  & $\RS$ & $\hsX$  & $\hsY$  \\
				\midrule
				1   & 0.731                                     & 6.837                                     & 7.063                                    & 0.893 & 4.219   & 4.113   & 0.914 & 4.553   & 4.394   \\
				2   & 0.798                                     & 4.704                                     & 4.538                                    & 0.728 & 6.257   & 5.679   & 0.573 & 6.672   & 6.719   \\
				3   & 0.831                                     & 6.185                                     & 6.392                                    & 0.824 & 4.715   & 5.039   & 0.870 & 4.649   & 4.849   \\
				4   & 0.867                                     & 4.757                                     & 5.020                                    & 0.838 & 5.250   & 5.355   & 0.813 & 4.901   & 5.404   \\
				5   & 0.753                                     & 6.798                                     & 5.310                                    & 0.795 & 5.597   & 4.920   & 0.853 & 5.055   & 4.807   \\
				6   & 0.730                                     & 5.624                                     & 4.631                                    & 0.694 & 8.054   & 5.540   & 0.817 & 5.421   & 4.276   \\
				7   & 0.796                                     & 5.872                                     & 4.848                                    & 0.844 & 5.509   & 4.660   & 0.877 & 6.895   & 5.020   \\
				8   & 0.805                                     & 4.865                                     & 5.473                                    & 0.798 & 5.715   & 5.010   & 0.825 & 5.640   & 4.006   \\
				9   & 0.771                                     & 5.668                                     & 5.681                                    & 0.729 & 5.472   & 6.538   & 0.873 & 5.328   & 4.914   \\
				10  & 0.786                                     & 5.111                                     & 6.125                                    & 0.878 & 4.430   & 5.348   & 0.896 & 5.342   & 6.132   \\
				11  & 0.231                                     & 8.709                                     & 272.743                                  & 0.359 & 5.824   & 298.676 & 0.012 & 21.137  & 702.646 \\
				12  & 0.019                                     & 690.530                                   & 181.928                                  & 0.002 & 396.639 & 415.174 & 0.004 & 579.639 & 332.651 \\
				\bottomrule
			\end{tabular}
		}
	\end{center}
	\caption{Results of fitting RPEs to a 2D Gaussian.}
	\label{tab:fitrelpos}
\end{table}

We also examined whether RPEs appear as 2D Gaussians across all layers (\figref{fig:relposbylayer}). We fitted the ERF of ViTs to a 2D Gaussian using the \texttt{LMfit} \citep{newville2016lmfit} library (\tabref{tab:fitrelpos}). The coefficient of determination $\RS$ indicates how exactly ERF fits a 2D Gaussian, ideally 1. The standard deviations $\hsX$ and $\hsY$ represent the wideness of the 2D Gaussian. We discovered that, for the majority of layers, RPE fitted to the 2D Gaussian with $\RS > 0.7$. The exceptions, whose RPE showed no pattern, were found in the last two layers of $l \in \{11, 12\}$ that were close to the classifier head.

We interpret these observations as follows: Initially, an untrained RPE exhibits a random pattern and cannot distinguish between near and far patches. However, because RPE is learnable, ViTs can choose to \emph{acquire} an understanding of the different positions of patches. After training, the RPE becomes a pattern close to a 2D Gaussian, discriminating the different positions of patches. The learned RPE allows ViTs to understand near and far patches. These observations motivate us to design a new RPE method.

\section{Proposed Method}
\label{sec:pro}
Our objective is to design an RPE that easily recognizes close and distant patches to facilitate ViTs to acquire spatial understanding of images. In light of the observation that learned RPE fits suitably with a 2D Gaussian, we propose injecting \emph{Gaussian attention bias} into RPE:
\begin{align}
	\Attention_l(\mathbf{Q}_l, \mathbf{K}_l, \mathbf{V}_l) = \softmax\left(\frac{\mathbf{Q}_l\mathbf{K}_l^\top}{\sqrt{D}} + \Brl +
	\begingroup\color{blue}\BGl\endgroup
	\right) \mathbf{V}_l.
\end{align}
Here, we aim to build $\BGl$ so that the bias terms $\Brl + \BGl$ readily appear as a 2D Gaussian, even in the initial state. By reversing the process of extracting RPE in \figref{fig:relposbypatch}, we build $\BGl$ by stacking sliced 2D Gaussians.

First, for the $l$th layer, we generate a 2D Gaussian table using $A_l, \sigma_l \in \real$:
\begin{align}
	f(x,\ y) & =A_l^2\exp\left(-\left(\frac{(x-x_c)^2}{2 \sigma_l^2} + \frac{(y-y_c)^2}{2 \sigma_l^2}\right)\right), \label{eq:gausstable}
\end{align}
where $x = 1, 2, \cdots, 2W/P-1$, $y = 1, 2, \cdots, 2H/P-1$, and $(x_c,\ y_c)$ correspond to the central coordinate. Note that the amplitude is set to $A_l^2$ to ensure a non-negative amplitude for any $A_l$. The variance $\sigma_l^2$ is shared for the horizontal and vertical directions. Thus, we parameterize the 2D Gaussian with only two parameters, $A_l$ and $\sigma_l$.

\begin{figure*}[t!]
	\begin{center}
        \includegraphics[width=0.99\linewidth]{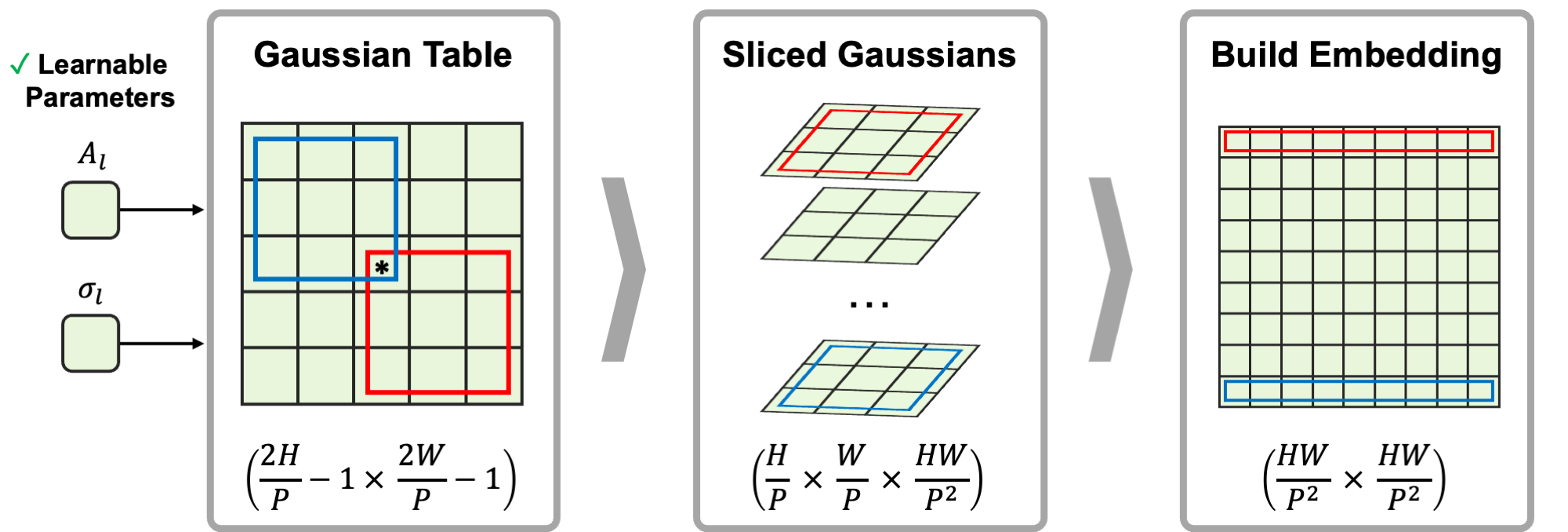}
	\end{center}
	\caption{Illustration on how we obtain Gaussian attention bias.}
	\label{fig:slicegaussian}
\end{figure*}

Second, sliced Gaussians are obtained. As shown in \figref{fig:slicegaussian}, the first sliced Gaussian should have the center coordinate of the 2D Gaussian (*) at the top-left (red box), whereas the last sliced Gaussian should exhibit the center coordinate (*) at the bottom-right (blue box). This slicing ensures that it resembles the learned RPE in \figref{fig:relposbypatch}. Finally, each sliced Gaussian is reshaped and stacked to build $\BGl$, whose size is the same as $\Brl$.

Our design of Gaussian attention bias has several advantages. First, because we designed Gaussian attention bias as an additional bias $\BGl$, it can be seamlessly plugged into any type of RPE, including RelPosBias and RelPosMlp. In other words, if we attempt to implement a redesign of RelPosMlp to resemble a 2D Gaussian, it cannot be applied to other RPEs, such as RelPosBias.

Second, our Gaussian attention bias is hyperparameter-free. Note that $\BGl$ is parameterized by $A_l$ and $\sigma_l$ with a differentiable function (\eqref{eq:gausstable}). Thus, $A_l$ and $\sigma_l$ can be set as learnable parameters in gradient descent optimization. We do not need the trial-and-error-based hyperparameter tuning on $A_l$ and $\sigma_l$. Because $\sigma_l$ determines the wideness of the 2D Gaussian, the flexibility of $\sigma_l$ is beneficial when using ViTs for other datasets or tasks that require different sizes of ERF. Furthermore, different $A_l$ and $\sigma_l$ values are allowed for each layer. As we observed in \tabref{tab:fitrelpos}, as the last two layers did not learn to be a 2D Gaussian, it is preferable to allow different behaviors in the last two layers. For example, the last two layers can naturally choose $A_l$ to be zero.

Finally, we benefit from the learnability of the original RPE, such as RelPosBias or RelPosMlp. Indeed, RPEs such as RelPosBias or RelPosMlp have a significant number of parameters that enable enriched expression in SA. Considering this behavior, we allow the degree of freedom of the original RPE.

However, we remove unnecessary degrees of freedom from our Gaussian attention bias. We do not generate multiple Gaussian tables; rather, we use a single Gaussian table to ensure that sliced Gaussians are shifted versions of each other, inspired by the use of relative coordinates in RPE. We do not use a constant term in our Gaussian function at \eqref{eq:gausstable} because $\softmax$ is invariant to constant translation \citep{DBLP:conf/nips/LahaCAKSR18,DBLP:conf/icml/MartinsA16}: $\softmax(\mathbf{x}+C) = \softmax(\mathbf{x})$. Finally, we choose to share the Gaussian attention bias across multiple heads of SA within the same layer (See the Appendix for the ablation study).

\section{Experiments}
\label{sec:exp}
Now, we investigate the influence of Gaussian attention bias on the performance of ViTs.

\begin{table}[t!]
	\begin{center}
		\begin{tabular}{ll|rrr}
			\toprule
			Dataset                      & Model                & RPE w/o GAB & RPE w/ GAB & Difference \\
			\midrule
			\multirow{3}{*}{ImageNet-1K} & ViT-S/16 (R) & 80.576      & 80.724     & +0.157     \\
			                             & ViT-M/16 (R) & 81.224      & 81.249     & +0.025     \\
			                             & ViT-B/16 (R) & 81.381      & 81.484     & +0.102     \\
			\bottomrule
		\end{tabular}
	\end{center}
	\caption{Top-1 accuracy on the ImageNet-1K dataset. All the accuracies in this paper are expressed in percentage units. ``GAB'' indicates Gaussian attention bias.}
	\label{tab:in}
\end{table}

\paragraph{ImageNet-1K} First, we trained the ViTs on the image classification task using the ImageNet-1K dataset \citep{DBLP:conf/cvpr/DengDSLL009} from scratch. ViT-$\{$S, M, L$\}$/16 (R) using RelPosMlp without APE were used. See the Appendix for experimental details, such as the hyperparameters. For each model with and without Gaussian attention bias, the top-1 accuracy was measured (\tabref{tab:in}). The top-1 accuracy of the three models improved after incorporating Gaussian attention bias.

\begin{table}[t!]
	\begin{center}
		\begin{tabular}{ll|rrr}
			\toprule
			Dataset                          & Model                & RPE w/o GAB & RPE w/ GAB & Difference \\
			\midrule
			\multirow{3}{*}{Oxford-IIIT Pet} & ViT-S/16 (R) & 91.486      & 92.780     & +1.294     \\
			                                 & ViT-M/16 (R) & 92.810      & 92.960     & +0.150     \\
			                                 & ViT-B/16 (R) & 93.381      & 93.743     & +0.362     \\
			\midrule
			\multirow{3}{*}{Caltech-101}     & ViT-S/16 (R) & 88.403      & 90.202     & +1.799     \\
			                                 & ViT-M/16 (R) & 89.132      & 89.983     & +0.851     \\
			                                 & ViT-B/16 (R) & 89.254      & 89.570     & +0.316     \\
			\midrule
			\multirow{3}{*}{Stanford Cars}   & ViT-S/16 (R) & 80.126      & 83.079     & +2.953     \\
			                                 & ViT-M/16 (R) & 80.731      & 83.890     & +3.159     \\
			                                 & ViT-B/16 (R) & 80.154      & 82.612     & +2.458     \\
			\midrule
			\multirow{3}{*}{Stanford Dogs}   & ViT-S/16 (R) & 81.535      & 82.507     & +0.972     \\
			                                 & ViT-M/16 (R) & 85.088      & 85.714     & +0.626     \\
			                                 & ViT-B/16 (R) & 89.256      & 90.185     & +0.929     \\
			\bottomrule
		\end{tabular}
	\end{center}
	\caption{Test accuracy with and without Gaussian attention bias on other datasets.}
	\label{tab:others}
\end{table}

\paragraph{Other Datasets} To further examine the performance difference, we targeted image classification on other datasets: Oxford-IIIT Pet \citep{DBLP:conf/cvpr/ParkhiVZJ12}, Caltech-101 \citep{DBLP:journals/cviu/Fei-FeiFP07}, Stanford Cars \citep{DBLP:conf/iccvw/Krause0DF13}, and Stanford Dogs \citep{khosla2011novel}. Test accuracy was measured for each ViT that used RelPosMlp with and without Gaussian attention bias. We observed that the use of Gaussian attention bias consistently improved the test accuracies of the three ViTs on the four datasets (\tabref{tab:others}). For these experiments, each dataset contained objects of various sizes, whose classification requires different sizes of ERF or $\sigma_l$. Because we designed $\sigma_l$ as learnable, the model with Gaussian attention bias can flexibly cope with different ERFs. Indeed, the learned $\sigma_l$ achieved similar but slightly different values for each dataset (See the Appendix).

\begin{table}[t!]
	\begin{center}
		\begin{tabular}{ll|ll|ll}
			\toprule
			\multirow{2}{*}{Backbone}       & \multirow{2}{*}{RPE Method} & \multicolumn{2}{c|}{COCO} & \multicolumn{2}{c}{ADE20K}                 \\
			                                &                             & AP$^{\textrm{box}}$       & AP$^{\textrm{mask}}$       & mIoU  & aAcc  \\
			\midrule
			\multirow{3}{*}{Swin-S} & RelPosBias w/o GAB          & 48.12                     & 43.03                      & 46.16 & 81.82 \\
			                                & RelPosBias w/ GAB           & 48.23                     & 43.13                      & 46.41 & 82.09 \\
			                                & Difference                  & +0.11                     & +0.10                      & +0.25 & +0.27 \\
			\bottomrule
		\end{tabular}
	\end{center}
	\caption{Experimental results in terms of object detection and semantic segmentation.}
	\label{tab:detseg}
\end{table}

\paragraph{Object Detection and Semantic Segmentation} Finally, we targeted two downstream tasks: object detection on the COCO 2017 dataset \citep{DBLP:conf/eccv/LinMBHPRDZ14} and semantic segmentation on the ADE20K dataset \citep{DBLP:journals/ijcv/ZhouZPXFBT19}. To further investigate our proposed method with a different setup, we targeted Swin-S with RelPosBias as the backbone. Using the COCO dataset, we measured bounding box mAP (AP$^{\textrm{box}}$) on object detection and segmentation mAP (AP$^{\textrm{mask}}$) on instance segmentation. Using the ADE20K dataset, the mean intersection over union (mIoU) and mean accuracy over all pixels (aAcc) were measured. We observed that the Swin transformers with Gaussian attention bias exhibited improvements across all four indices (\tabref{tab:detseg}).

\section{Conclusion}
\label{sec:con}
In this study, we analyzed how ViTs understand spatial images. From detailed analyses of the ERF of ViTs, we discovered that ViTs acquired spatial understanding of images during training and that this phenomenon was caused by the underlying transition from randomized positional embedding to a learned one. To guide the understanding of the spatial nearness and farness of patches, we proposed injecting Gaussian attention bias into ViTs. In several experiments on image classification, object detection, and semantic segmentation, ViTs integrated with Gaussian attention bias achieved superior results.

\bibliographystyle{unsrt}
\bibliography{egbib}
\end{document}